\documentclass{article}

\PassOptionsToPackage{numbers, compress}{natbib}
\usepackage[sglblindworkshop, final]{neurips_2025}

\usepackage[utf8]{inputenc} 
\usepackage[T1]{fontenc}    
\usepackage{hyperref}       
\usepackage{url}            
\usepackage{booktabs}       
\usepackage{amsfonts}       
\usepackage{nicefrac}       
\usepackage{microtype}      
\usepackage{xcolor}         
\usepackage{multirow}
\usepackage{amsmath}
\usepackage{booktabs} 
\usepackage[ruled,vlined,linesnumbered,longend]{algorithm2e}
\usepackage{graphicx}
\usepackage{wrapfig}
\usepackage{caption}
\usepackage[most]{tcolorbox}
\usepackage{xcolor} 

\tcbset{
  promptbox/.style={
    colback=gray!10,    
    colframe=gray!80,   
    fonttitle=\bfseries,
    title=LLM Prompt,   
    boxrule=0.5pt,      
    arc=3pt,            
    outer arc=3pt,
    top=4pt, bottom=4pt,
    left=6pt, right=6pt
  }}

\title{Weakly-supervised Latent Models for Task-specific Visual-Language Control}
\workshoptitle{
Space in Vision, Language, and Embodied AI}

\author{%
  Xian Yeow Lee \\
  Industrial AI Lab, Hitachi America, Ltd. \\
  \texttt{xian.lee@hal.hitachi.com}
  \And
  Lasitha Vidyaratne \\
  Industrial AI Lab, Hitachi America, Ltd. \\
  \texttt{lasitha.vidyaratne@hal.hitachi.com}
  \And
  Gregory Sin \\
  Industrial AI Lab, Hitachi America, Ltd. \\
  \texttt{gregory.sin@hal.hitachi.com}
  \And
  Ahmed Farahat \\
  Industrial AI Lab, Hitachi America, Ltd. \\
  \texttt{ahmed.farahat@hal.hitachi.com}
  \And
  Chetan Gupta \\
  Industrial AI Lab, Hitachi America, Ltd. \\
  \texttt{chetan.gupta@hal.hitachi.com}
}

\begin{document}

\maketitle

\begin{abstract}
Autonomous inspection in hazardous environments requires AI agents that can interpret high-level goals and execute precise control. A key capability for such agents is spatial grounding, for example when a drone must center a detected object in its camera view to enable reliable inspection. While large language models provide a natural interface for specifying goals, using them directly for visual control achieves only 58\% success in this task. We envision that equipping agents with a world model as a tool would allow them to roll out candidate actions and perform better in spatially grounded settings, but conventional world models are data and compute intensive. To address this, we propose a task-specific latent dynamics model that learns state-specific action-induced shifts in a shared latent space using only goal-state supervision. The model leverages global action embeddings and complementary training losses to stabilize learning. In experiments, our approach achieves 71\% success and generalizes  to unseen images and instructions, highlighting the potential of compact, domain-specific latent dynamics models for spatial alignment in autonomous inspection.

\end{abstract}

\section{Introduction}
Autonomous inspection systems offer deployment potential in industrial environments where hazards such as high-altitude structures, confined spaces, or toxic atmospheres make human operation unsafe~\cite{nooralishahi2021drone, jenssen2018automatic, rakha2018review}. These systems must integrate perception, task understanding, and precise control, capabilities well suited to AI agent architectures. World models improve planning and sample efficiency, while language models provide a natural interface for operators to specify high-level goals and convert them into executable actions~\cite{zhou2024dino, Wu2022DayDreamerWM, Hafner2020MasteringAW, Sekar2020PlanningTE, zhang2023storm}. A common use case involves autonomous or semi-autonomous inspection with a human in the loop. An operator might issue a command such as "Inspect for defects under the bridge," which the agent interprets, plans, optionally simulates using a world model, and executes the best action. This agentic approach reduces workload, speeds positioning, increases safety and enables more flexible and adaptive inspection strategies compared to rigid, predefined workflows.

While full-fidelity world models could allow agents to simulate many possible action sequences, they are data and compute intensive and often unnecessary for targeted tasks~\cite{fu2024exploring, Cong2025CanTSA, Baldassarre2025BackTTA}. We propose a task-specific latent dynamics model that captures only the action-conditioned spatial shifts in the latent space required for centering the object, enabling effective planning with limited supervision. We focus on a subproblem of visual inspection: the agent has detected the target object in its camera view, but it is off-center. The task is to generate motion commands that bring the object into the center of the images, testing spatially grounded planning and requiring precise visual-motor coupling.

\begin{wrapfigure}{r}{0.49\textwidth}
  \begin{center}
    \includegraphics[width=0.48\textwidth]{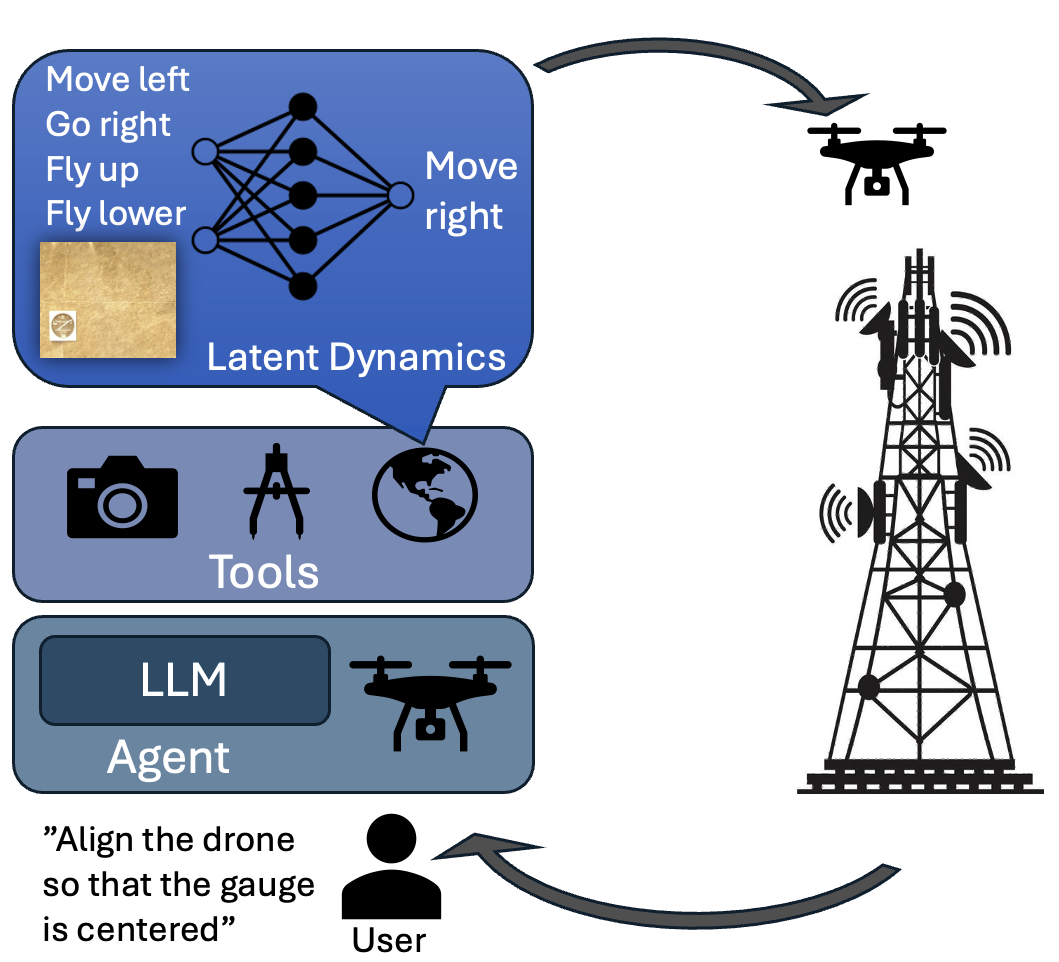}
  \end{center}
  \caption{Overview of an LLM-powered AI agent leveraging tools, including a latent dynamics model, for improved spatially grounded visual-language control in inspection tasks.}
\label{fig:overview}
\end{wrapfigure}

This workshop paper's contributions are as follows: First, we identify a scenario where naive multimodal LLM-based planning achieves limited success, even with careful prompting and reasoning models. Second, we propose a latent dynamics model specialized for object centering. Thirdly, we show that the model can roll out effective actions using only goal-state supervision and demonstrate that compact, task-specific latent dynamics models trained with limited data can outperform larger foundational models for spatial alignment tasks. Figure~\ref{fig:overview} provides an overview of our LLM-powered AI agent using a latent dynamics model for this inspection task.

\section{Related Works}

Research at the intersection of language, vision, and action has explored how agents interpret natural language instructions and interact with the environment. Early benchmarks such as ALFRED introduced household-scale instruction-following tasks \cite{shridhar2020alfred}, while subsequent work integrated large language models with embodied agents, enabling robots to sequence skills from high-level commands \cite{ahn2022can}. More recent efforts, including CALVIN \cite{mees2022calvin} and MineDojo \cite{fan2022minedojo}, provide long-horizon, multimodal benchmarks testing agents’ ability to ground instructions in visual and interactive contexts.

Visual-language models (VLMs) have shown strong generalization for object identification, spatial queries, and zero-shot navigation, though most evaluations emphasize semantic rather than fine-grained control. Models such as LLaVA \cite{liu2023visual} and Flamingo \cite{alayrac2022flamingo} demonstrate multimodal reasoning in dialogue, but targeted studies reveal VLMs underperform on precise spatial reasoning tasks \cite{chen2024spatialvlm,wang2024picture}.

Model-based approaches using learned world models have improved planning, sample efficiency, and generalization in both simulated and real-world settings \cite{hafner2023mastering}. Methods like DreamerV3 leverage compact latent dynamics for scalable, data-efficient reinforcement learning, and latent predictive learning frameworks such as JEPA \cite{lecun2022path} show that abstract representations improve generalization in data-limited regimes. However, these approaches often face high computational cost, sparse supervision, and difficulty capturing precise action semantics for fine-grained control. In contrast, our work focuses on \emph{task-specific dynamics modeling}, learning only the dynamics needed to achieve a target goal with limited supervision, using initial and goal states rather than full trajectories \cite{finn2017one,ebert2018visual,nair2022r3m}. By operating in a latent space, we capture explicit action semantics without sequential action data, enabling data-efficient planning for centering objects.

Alternative spatially grounded control methods include traditional vision-based pipelines that compute object displacements with camera calibration and bounding box detection \cite{corke2011robotics}. While effective, these require engineering effort, careful calibration, and requires further integration with language models. Zero-shot multi-modal LLM planning offers a data-efficient alternative~\cite{Wong2024AffordancesOrientedPUA}, but our experiments (Section~\ref{subsec:llm_planners}) show limited performance for precise, visually grounded tasks. Table~\ref{tab:related} provides a qualitative comparison of zero-shot LLM planning, conventional world models, and our latent dynamics model, highlighting the importance of explicit action semantics for spatial alignment.

\begin{table}[h]
\small
\centering
\begin{tabular}{lcccc}
\toprule
     & Sequential & Action  & Task  & Compute\\
Method &  Data &  Representation & Specific &  Cost \\

\midrule
Zero-shot LLM Planning & No & Implicit & No & Low \\
Conventional World Model & Yes & Explicit (Observation) & No & High \\
Latent Dynamics Model (this work) & No & Explicit (Latent) & Yes & Low \\
\bottomrule
\end{tabular}
\caption{Comparison of approaches highlighting data requirements, how actions are represented, task specificity, and computational cost.}
\label{tab:related}
\end{table}

\section{Problem Formulation and Method}

\subsection{Problem Definition}
We consider the problem of training a task-specific latent dynamics model for autonomous alignment in drone-based inspection since prior studies have shown that learning a dynamics model in a latent space can yield better generalization than direct action prediction, especially when data quality is limited~\cite{sobal2025learning}. An agent (a drone) receives a natural-language instruction (e.g., \textit{"center the object"}) and must select discrete movement actions to center a target object in the frame. We focus on this problem because centering is the minimal subproblem requiring vision–language–action grounding and precise control. Success here is a strong indicator for generalization to more complex inspection tasks. A state $s$ is defined as the tuple $(x_{\text{img}}, x_{\text{instr}})$, where $x_{\text{img}}$ is the current off-center image and $x_{\text{instr}}$ is the textual instruction. The goal state $s^*$ is any image where the object of interest (e.g., gauges, switches, or structural features) is centered. For practical applications, we assume a small set of goal images is available, which can be safely captured in advance. Our dataset consists of $(s, a, s^*)$ triples, where $a \in \mathcal{A}$ is the correct action vector that will lead the drone towards the $s^*$ (e.g., \texttt{left}, \texttt{right}, \texttt{up}, \texttt{down}, \texttt{none}). Importantly, intermediate states $s_{t+1}$ for arbitrary actions are \emph{not} available, which prevents standard supervised transition-model learning. 

\subsection{Latent Dynamics Model}

In our formulation, both states and actions are embedded into a shared $d$-dimensional latent space $\mathcal{Z}$ using separate image, instruction and action encoders $E^{img}_\phi$, $E^{inst}_\phi$, $E^{act}_\phi$:
\begin{equation}
z_s = \mathrm{concat(}E^{img}_\phi(x_{\text{img}}), E^{inst}_\phi(x_{\text{instr}})) \in \mathcal{Z}, \quad 
z_a = E_\phi^{\text{act}}(a) \in \mathcal{Z}.
\end{equation}

Goal embeddings are obtained by encoding the available goal images into latent space and averaging to obtain a goal prototype, $z^*$:
\begin{equation}
z^* = \frac{1}{N} \sum_{i=1}^N E^{img}_\phi(s^*_{\text{i}})
\end{equation}

Predicting the next state directly in latent space can lead to degenerate solutions where the model ignores action semantics and outputs the goal embedding for all inputs. As such, the latent dynamics model $f_\theta$ is instead trained to predict $\Delta_\theta$, where $\Delta_\theta$ is the state-specific action-induced shift in the latent embedding space. 
\begin{equation}
\Delta_\theta(z_s, z_a) = f_\theta(z_s, z_a)
\end{equation}
and the state transition is approximated by
\begin{equation}
\hat{z}_{s^{*}} \approx \hat{z_{s^{'}}} =  z_s + \Delta_\theta(z_s, z_a),
\end{equation}
Note that the above approximation is not strictly accurate. Our collected action data does not guarantee that the action will move the drone immediately to the goal state; it only ensures that the action directs the drone toward the goal. Since the ground truth of the next state is unavailable under our assumptions, the latent dynamics model must be trained using only weak supervision from the goal state prototypes, together with additional constraints, as described in the next section.

\subsection{Training Objectives}
To ensure that the latent dynamics model is robust and does not learn spurious shortcut transitions in the latent space, we train the model using four complementary losses:
\begin{equation}
\mathcal{L}_{\text{total}} =
w_{\text{dir}}\mathcal{L}_{\text{dir}} + 
w_{\text{rank}}\mathcal{L}_{\text{rank}} +
w_{\text{cons}}\mathcal{L}_{\text{cons}} +
w_{\text{reg}}\mathcal{L}_{\text{reg}}.
\end{equation}

\textbf{Directional and Ranking Losses:} These losses ensure that the predicted next state moves closer to the goal, and that the correct action ranks highest among all candidate actions:
\begin{align}
\mathcal{L}_{\text{dir}} &= \max\big(0, D(\hat{z}_{s'}, z^*) - D(z_s, z^*) + m\big), \\
\mathcal{L}_{\text{rank}} &= \mathrm{CE}\Big(\mathrm{softmax}\big(-D(\hat{z}_{s'}^{(i)}, z^*)/\tau\big), y\Big),
\end{align}
where $\hat{z}_{s'}^{(i)}$ is the predicted next-state embedding for action $a_i$, $\tau$ is a temperature parameter, $y$ is the index of the correct action, $m$ is the margin hyperparameter and $D(\cdot,\cdot)$ is a distance metric (e.g., Cosine similarity, or Euclidean distance). The directional loss $\mathcal{L}_{\text{dir}}$ encourages the model to move the state embedding closer to the goal prototype $z^*$, while the ranking loss $\mathcal{L}_{\text{rank}}$ ensures that the correct action is preferred over alternative actions.

Additionally, we introduce a \emph{global} action embedding, $g_a$, for each discrete action $a \in \mathcal{A}$, which is learned simultaneously with the latent dynamics model. These vectors represent the typical effect of an action across all states and serve as stable reference points in the latent space. For example, the global shift for the action \texttt{left} encodes the general tendency to move the object right in the image, independent of the current state. The state-specific delta $\Delta_\theta(z_s, z_a)$ then adapts this global shift to the current state, capturing situational variations such as object position or perspective. Global shifts are used only during training to guide and stabilize learning through the consistency and regularization losses and are not required at inference time. To our knowledge, global action embeddings have not been explicitly used to stabilize weakly supervised latent dynamics training.

\textbf{Consistency and Regularization Losses:} With the global action embeddings, we use the following losses to further promote stable and consistent action semantics in the latent space:
\begin{align}
\mathcal{L}_{\text{cons}} &= \| \Delta_\theta(z_s, z_a) - g_a \|_2^2, \\
\mathcal{L}_{\text{reg}}  &= \| g_a \|_2^2.
\end{align}
The consistency loss $\mathcal{L}_{\text{cons}}$ aligns the state-specific predicted shifts $\Delta_\theta(z_s, z_a)$ with the corresponding global action embeddings $g_a$, ensuring that the model produces coherent and semantically meaningful movements across different states. The regularization loss $\mathcal{L}_{\text{reg}}$ prevents the global shifts from growing excessively large, which could destabilize training or produce unrealistic latent transitions.

\subsection{Model architectures and training}

Algorithm~\ref{alg:train} summarizes the end-to-end training procedure of the model, and we use the following architectures for each component in our proposed model:

\begin{algorithm}
\small
\caption{Latent Dynamics Model Training}
\label{alg:train}
\KwIn{Training dataset $\mathcal{D}$, goal prototype $z^*$, learning rate $\eta$, loss weights $\mathbf{w}$, temperature $\tau$, margin $m$}
\KwOut{Trained model parameters $\{\phi_{\text{img}}, \phi_{\text{inst}}, \phi_{\text{act}}, \theta, \mathbf{g}_a\}$}
Initialize model parameters and optimizers\;
\For{epoch = 1 to $N_{\text{epochs}}$}{
    \For{batch $\mathcal{B} \in \mathcal{D}$}{
        $\text{Encode images: } \mathbf{z}_{\text{img}} \gets E^{\text{img}}_{\phi}(x_{\text{img}})$\;
        $\text{Encode instructions: } \mathbf{z}_{\text{inst}} \gets E^{\text{inst}}_{\phi}(x_{\text{instr}})$\;
        $\text{Concatenate state: } \mathbf{z}_s \gets \text{concat}(\mathbf{z}_{\text{img}}, \mathbf{z}_{\text{inst}})$\;
        
        \ForEach{action $a$ in batch}{
            $\text{Encode action: } \mathbf{z}_a \gets E^{\text{act}}_{\phi}(a)$\;
            $\text{Predict delta: } \Delta_\theta \gets f_\theta(\mathbf{z}_s, \mathbf{z}_a)$\;
            $\text{Compute next state: } \hat{\mathbf{z}}_{s'} \gets \mathbf{z}_s + \Delta_\theta$\;
            $\text{Directional loss: } 
            \mathcal{L}_{\text{dir}} \gets \max(0, D(\hat{\mathbf{z}}_{s'}, z^*) - D(\mathbf{z}_s, z^*) + m)$\;
            
            \ForEach{action $a_i \in \mathcal{A}$}{
                $\Delta_\theta^{(i)} \gets f_\theta(\mathbf{z}_s, E^{\text{act}}_{\phi}(a_i))$\;
                $\hat{\mathbf{z}}_{s'}^{(i)} \gets \mathbf{z}_s + \Delta_\theta^{(i)}$\;
                $d_i \gets D(\hat{\mathbf{z}}_{s'}^{(i)}, z^*)$\;
            }
            
            $\text{Ranking loss: } 
            \mathcal{L}_{\text{rank}} \gets \text{CrossEntropy}(\text{softmax}(-\mathbf{d}/\tau), y_{\text{true}})$\;
            
            $\text{Consistency loss: } \mathcal{L}_{\text{cons}} \gets \|\Delta_\theta - \mathbf{g}_a\|_2^2$\;
            $\text{Regularization loss: } \mathcal{L}_{\text{reg}} \gets \|\mathbf{g}_a\|_2^2$\;
        }
        
        $\mathcal{L}_{\text{total}} \gets w_{\text{dir}}\mathcal{L}_{\text{dir}} 
        + w_{\text{rank}}\mathcal{L}_{\text{rank}} 
        + w_{\text{cons}}\mathcal{L}_{\text{cons}} 
        + w_{\text{reg}}\mathcal{L}_{\text{reg}}$\;
        
        $\text{Update parameters: } 
        \{\phi, \theta, \mathbf{g}_a\} \gets 
        \{\phi, \theta, \mathbf{g}_a\} 
        - \eta \nabla_{\{\phi, \theta, \mathbf{g}_a\}}\mathcal{L}_{\text{total}}$\;
    }
}
\end{algorithm}

\textbf{Encoders}: Inputs are mapped into a shared $d$-dimensional latent space via three separate encoders. The image encoder $E^{\text{img}}_{\phi}$ utilizes a ResNet architecture with residual connections. The instruction encoder $E^{\text{inst}}_{\phi}$ is a lightweight transformer with positional encodings and a transformer encoder. The action encoder $E^{\text{act}}_{\phi}$ is realized by maintaining separate embedding tables for each discrete action.

\textbf{Dynamics Model}: The dynamics model $f_\theta$ is implemented as a multilayer perceptron that predicts state-specific action-induced shifts $\Delta_\theta$ in the latent space. The network processes concatenated state and action embeddings and outputs the predicted transition vector.

\textbf{Global Action Embeddings}: Global action embeddings $\mathbf{g}a$ are implemented as learnable parameters representing the typical behavior of each discrete action $a \in \mathcal{A}$ in the latent space. 

Once the model is trained, the latent dynamics model can then be used for planning in latent space. At each step, given an image and a textual instruction, the model predicts the next-state embeddings for all actions and selects the action minimizing distance to the goal prototype. Execution continues until the object is centered, defined as being within $\epsilon$ pixels of the image center. Algorithm~\ref{alg:test} in the Appendix summarizes this process.

\section{Results}
\subsection{Multi-modal LLM as Planners}\label{subsec:llm_planners}

\begin{figure}[h!]
\centering 
\includegraphics[width=0.85\linewidth]{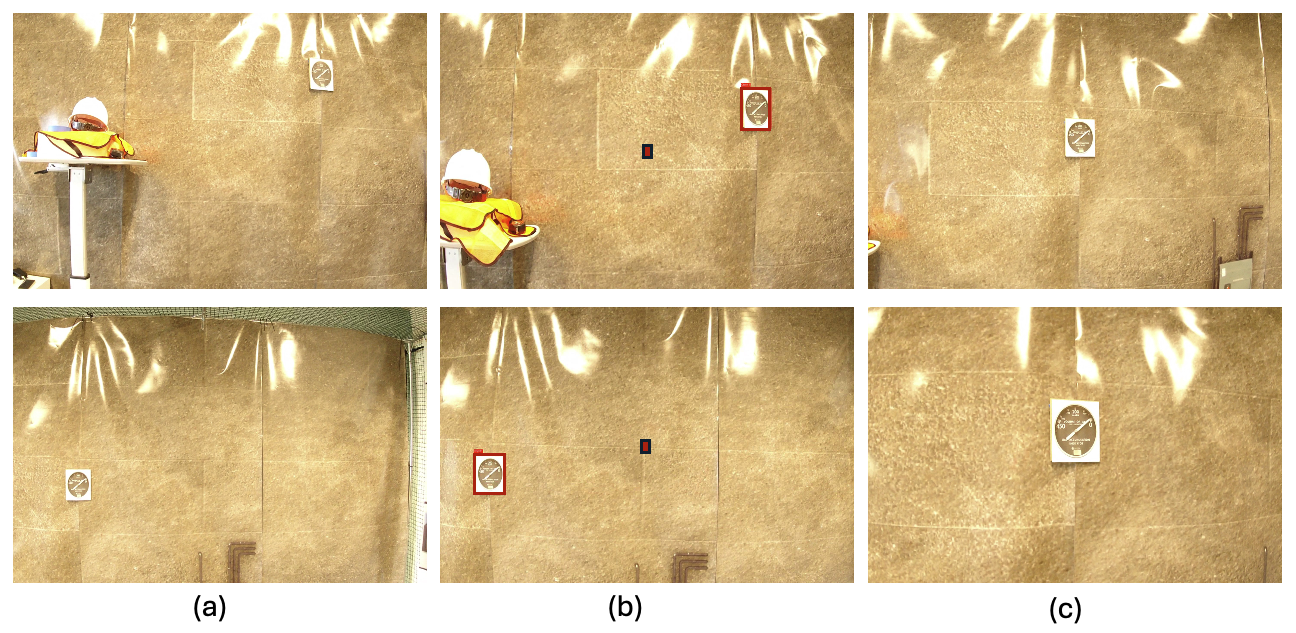} 
\caption{ Example images of a pressure gauge. (a) Raw images, (b) Images with bounding box annotations, (c) Ideal images with gauge centered.} 
\label{fig:images} 
\end{figure}

\begin{table}[ht]
\centering
\small
\begin{tabular}{l c c c c}
\toprule
\textbf{Input} & \textbf{Mode} & \textbf{Instruction} & \textbf{Acc. (\%) -Gemini} & \textbf{Acc. (\%) - GPT} \\
\midrule
Image + Annotations & R  & Varied   & 26.5 $\pm$ 4.9 & 57.0 $\pm$ 4.8 \\
Image + Annotations & R  & Fixed    & 36.0 $\pm$ 4.2 & 58.0 $\pm$ 4.5 \\
Image + Annotations & NR & Varied   & 38.5 $\pm$ 2.9 & 33.5 $\pm$ 6.3 \\
Image + Annotations & NR & Fixed    & 42.5 $\pm$ 3.1 & 34.5 $\pm$ 2.7 \\
\hline
Raw Image           & R  & Varied   & 14.0 $\pm$ 4.2 & 44.5 $\pm$ 5.1\\
Raw Image           & R  & Fixed    & 9.5  $\pm$ 3.3 & 48.0 $\pm$ 5.4 \\
Raw Image           & NR & Varied   & 29.0 $\pm$ 9.6 & 21.5 $\pm$ 5.5 \\
Raw Image           & NR & Fixed    & 34.0 $\pm$ 1.4 & 29.5 $\pm$ 3.7 \\
\bottomrule
\end{tabular}
\caption{\small Performance across input type, reasoning (R) vs non-reasoning (NR), and instruction type -varied vs. fixed across 5 random trials.}
\label{tab:llm_zeroshot}
\end{table}

We first evaluate whether multimodal LLMs has sufficient spatial knowledge to act as a spatial planner. Given an image and a textual instruction to center the object of interest in the image (detailed prompt shown in Appendix~\ref{app:prompt}), the LLM is tasked to output a two-dimensional direction vector selected from a discrete set of movement commands: left, right, up, down, or none. This vector corresponds to the motion the drone should take. To test this, we collected 40 test images of a pressure gauge in a mock industrial setting. The images were taken by a drone from various positions, as shown in Figure~\ref{fig:images}(a). As a simplification, we treat this as a two-dimensional control problem with fixed depth, although the images were captured from different distances from the gauge.

For the experiments, we use two relatively frontier models at the time of writing, Gemini-2.5 Flash in both reasoning and non-reasoning modes as well as GPT-4.1 for non-reasoning mode and o3 for reasoning mode. We also perform bounding box detection on the object of interest and annotate it in the image, as shown in Figure~\ref{fig:images}(b), to test whether a more explicit representation improves performance. In addition, we study the effect of prompt variation. For experiments with fixed prompts, we use a single manually defined prompt for all 40 images. For experiments with varying prompts, we use the same manually defined prompt and generate 40 additional variations with an LLM. Table~\ref{tab:llm_zeroshot} shows that bounding box information provides slight improvements, and prompt variation has only minor effects though overall performance remains low. Even with explicit bounding boxes and reasoning prompts, accuracy remains <58\%, showing lack of grounded spatial control. This suggests that while LLMs can interpret relative positions, they lack a grounded world model that connects actions to visual outcomes.

\subsection{Latent Dynamics Model as a Planner}

\textbf{Data collection and Experiments}

To train the proposed latent model, we collected a relatively small amount of additional training data. Recall that training requires access to samples of the goal state as well as samples of various initial states. We collected 100 images in which the gauge was approximately centered and encode them using Gemini as a trained image embedder. For initial states, we divided the space around the gauge into eight discrete quadrants (north, northeast, east, and so on) and programmed the drone to fly to random locations within each quadrant. The drone was positioned at different depths and orientations to capture diverse images with variation in yaw and distance from the gauge. This setup provided a straightforward way to generate correct action direction labels for each image. We collected approximately 200 samples from each quadrant. Each image was paired with either a fixed instruction or a varied instruction to form the dataset. Training followed the procedure in Algorithm~\ref{alg:train}. To further enhance the dataset, we applied standard non-geometric image augmentation techniques to each batch in order to improve generalization while preserving geometry so that the corresponding action labels remained valid.

We conducted experiments with both fixed and varying instructions. We also compared different distance metrics, including cosine similarity, Euclidean distance, and a combination of both, to evaluate their effect on performance. All experiments were conducted on a single consumer-grade RTX-4090 GPU and trained for 50 epochs using five random seeds. During training, we monitored validation accuracy and saved the best-performing model checkpoint for final evaluation. The dataset was split into 80 percent training and 20 percent validation, with the 40 images collected in the previous section reserved as a fixed held-out test set. A complete list of experimental hyperparameters is provided in the Appendix.

\textbf{Results}

Table~\ref{tab:latent_model_combined} presents the results of using the trained latent dynamics model to plan the next best action to center the object, evaluated on the held-out test images under two conditions. The first evaluation pairs the images with instructions drawn from the training set, randomly matched so that specific image–instruction pairs differ from those seen in training. The second evaluation instead uses newly generated instructions not included in training, while keeping the same held-out images. The first setting therefore measures generalization primarily in the visual domain, while the second measures generalization across both vision and language.

Several observations can be made. First, training a domain-specific latent dynamics model for planning consistently outperforms directly using a multimodal LLM for planning, even when varying hyperparameters. Second, the type of instruction, whether fixed or varied, does not lead to large differences in accuracy, suggesting robustness of the model to instruction style. Third, models trained with a cosine similarity component generally achieve higher accuracy than those trained solely with Euclidean distance. Across both evaluations, the combination of cosine similarity and Euclidean distance does not yield a consistent advantage over cosine similarity alone.

The best model achieves an accuracy of 70.5\% in the first evaluation and 71.0\% in the second evaluation. Importantly, performance in the second setting does not degrade compared to the first, indicating that the latent model generalizes well to previously unseen images and instructions drawn from similar distributions. Both results are significantly higher than the multimodal LLM baselines, which achieves 48\% with raw images and 58\% with annotated images, which also requires an additional detection module and is based on the assumption that the annotations are accurate. These findings highlight that the proposed latent dynamics model generalizes effectively within the task distribution while maintaining robustness across both visual and textual variation.

\begin{table}[ht]
\small
\centering
\begin{tabular}{c|c|c|c}
\hline
 & & \textbf{Accuracy } & \textbf{Accuracy} \\
\textbf{Instruction} & \textbf{Distance}  & \textbf{(Vision Gen.)} & \textbf{(Vision + Text Gen.)} \\
\hline
Fixed & Cosine Similarity & 70.5 $\pm$ 7.6 & 71.0 $\pm$ 6.9 \\
Fixed & Euclidean & 67.5 $\pm$ 7.7 & 68.0 $\pm$ 7.4 \\
Fixed & Cosine Similarity + Euclidean & 69.5 $\pm$ 6.7 & 70.5 $\pm$ 5.4 \\
Varied & Cosine Similarity & 70.0 $\pm$ 6.1 & 70.0 $\pm$ 6.1 \\
Varied & Euclidean & 68.5 $\pm$ 8.0 & 69.5 $\pm$ 7.8 \\
Varied & Cosine Similarity + Euclidean & 70.5 $\pm$ 4.8 & 70.5 $\pm$ 4.8 \\
\hline
\end{tabular}
\caption{Performance of the latent dynamics model on held-out test images. The first evaluation uses training-style instructions (vision generalization), while the second uses newly generated instructions (vision and text generalization).}
\label{tab:latent_model_combined}
\end{table}

\subsection{Ablation Studies}

Next, we perform an ablation study to determine which components contribute most to overall performance. We use cosine similarity as the distance metric with varied instructions, as this corresponds to the best distance hyperparameter and represents a typical use case where users provide semantically similar but different instructions. Similar to previous experiments, for each ablation study we trained 5 seeds and evaluated the models on the set of unseen images and instructions. For each experiment, we systematically remove one or more loss function components: the directional loss (\(\mathcal{L}_{\text{dir}}\)), the ranking loss (\(\mathcal{L}_{\text{rank}}\)) and the consistency loss (\(\mathcal{L}_{\text{cons}}\)) together with the regularization loss (\(\mathcal{L}_{\text{reg}}\)) that operates on the global action embeddings, to measure their individual contributions.  

Table~\ref{tab:ablation} shows the impact of each ablation on the accuracy. Interestingly, removing the directional loss results in a slight increase in performance to 72.0\% ± 3.3\%, slightly higher than the baseline of 71.0\% ± 6.9\%, indicating that this component is not critical in this setting. We hypothesize that the directional loss may over-penalize ambiguous near-center states, making ranking loss alone more effective. Removing the consistency and regularization losses leads to a moderate decrease to 68.5\% ± 4.2\%. In contrast, removing the ranking loss causes a dramatic drop to 12.0\% ± 9.1\%, demonstrating that this component is essential for the model to perform well. These results highlight that while all components contribute, the ranking loss is the most crucial for guiding the model, with consistency and directional losses providing smaller but still beneficial effects.

\begin{table}[ht]
\small
\centering
\begin{tabular}{c|c}
\hline
\textbf{Ablation Type} & \textbf{Rank-1 Accuracy} \\
\hline
No Consistency/Regularization & 68.5\% ± 4.2\% \\
No Directional Loss            & 72.0\% ± 3.3\% \\
No Ranking Loss                & 12.0\% ± 9.1\% \\
\hline
\end{tabular}
\caption{Ablation study evaluating the contribution of each loss component.}
\label{tab:ablation}
\end{table}

\subsection{Analysis of failures}
\begin{figure}[h]
    \centering
    \includegraphics[width=0.9\linewidth]{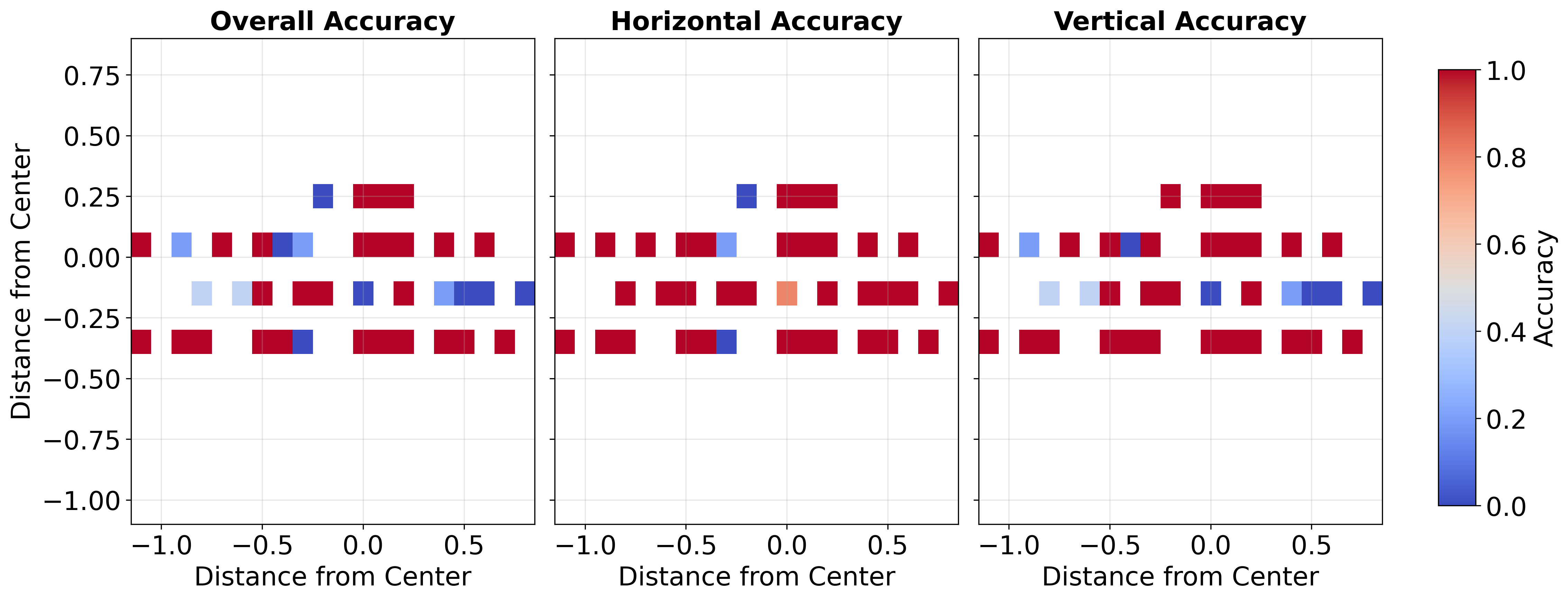}
    \caption{Heatmaps of overall, horizontal, and vertical accuracy showing that most errors occur when the drone is near the image centerline.}
    \label{fig:analysis}
\end{figure}

\begin{figure}[h!]
\centering 
\includegraphics[width=0.75\linewidth]{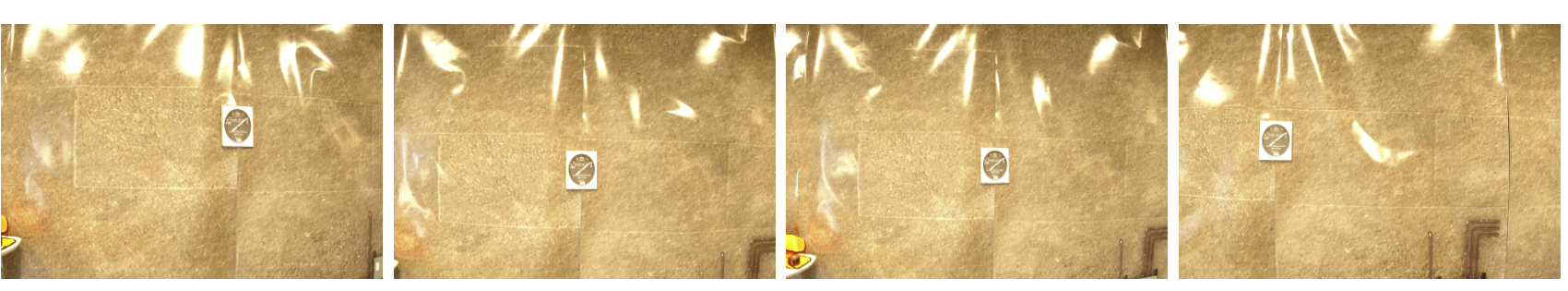} 
\caption{Sample images where the latent model fails to estimate the best actions to center the object in the image.} 
\label{fig:failures} 
\end{figure}

We analyze model failures on the test set by visualizing the drone’s position relative to the image center and plotting heatmaps of average accuracy across all random seeds at those locations. Figure~\ref{fig:analysis} shows heatmaps for overall accuracy, horizontal control accuracy, and vertical control accuracy. When decomposing overall accuracy into horizontal and vertical components, we observe that most failures occur when the drone is positioned near the horizontal or vertical centerline. In these cases, the model is often uncertain whether a corrective movement is necessary, leading to misclassifications of the action as either a shift or no movement. This suggests that the model’s main source of error lies in fine-grained decision making near the center, where distinctions between valid actions are most subtle. Figure~\ref{fig:failures} shows some of the representative samples of failure cases.



\subsection{Benefits and Limitations}
Our approach provides practical advantages for real-world tasks, particularly in scenarios where collecting sequential interaction data is difficult or unsafe. The method requires only random initial states and goal images for training, making data collection feasible and safe since both centered and off-centered states can be scripted. It generalizes across unseen images and diverse natural language instructions, and consistently outperforms multi-modal LLMs in the task, demonstrating that lightweight, task-specific latent models can offer an efficient and effective solution for grounded spatial control.

At the same time, the proposed model does not constitute a general world model. It is designed for fixed-goal tasks and does not capture full environmental dynamics, stochastic behaviors, or all possible transitions, which limits its applicability to open-ended or multi-step planning scenarios, which we plan to extend upon in future works. The method also depends on the quality of latent representations, and its robustness under more complex objects or environments has not yet been evaluated. Despite these limitations, we believe this approach strikes a practical balance between simplicity, safety, and performance for goal-directed spatial planning.

\section{Conclusion}

We present a latent dynamics model for goal-directed visual planning, achieving over 70\% accuracy on unseen images and instructions, outperforming zero-shot LLM planners. Task-specific latent modeling provides reliable, grounded action selection, complementing agentic frameworks as a tool. While limited to single-object centering and a fixed action space, our approach highlights the value of ranking-based losses and structured latent representations. Future work includes scaling the approach from 2D to 3D to include depth, and more objects of interest, testing transferability from simulation to real environments, and incorporating negative states to capture failure modes. Additionally, we plan to leverage LLMs to generate candidate plans and evaluate them efficiently using the lightweight dynamic model, rather than relying on brute-force planning or pure LLM reasoning for more unstructured free-form actions. Overall, lightweight latent models offer a promising route for grounded perception as a tool for agentic systems that operates in the physical world.

\section{Broader Impacts}
This work presents a method for training models that enable agentic systems to control autonomous platforms and center objects of interest within images. While the method itself is not inherently harmful, we acknowledge its potential misuse in applications such as large-scale surveillance, and thus recognize the importance of considering dual-use risks. However, in the context of automated inspection, this approach can provide substantial benefits by reducing the physical risks faced by workers in hazardous environments, improving the accuracy and efficiency of inspection task, and enhancing the overall reliability of safety assessments. These outcomes can contribute both to improved worker safety and to greater scalability and cost-effectiveness in industrial inspection workflows.
\bibliographystyle{plainnat} 
\bibliography{references}    

\appendix

\section{Appendix}
\subsection{Using the Latent Dynamics during Model Test Time}\label{app:alg}

\begin{algorithm}[H]
\small
\label{alg:test}
\caption{Planning with Latent Dynamics Model}
\KwIn{Current image and instruction $(x_{\text{img}}, x_{\text{instr}})$, goal prototype $z^*$, action set $\mathcal{A}$, encoders $E^{\text{img}}_{\phi}, E^{\text{inst}}_{\phi}, E^{\text{act}}_{\phi}$, latent dynamics model $f_\theta$}
\KwOut{Actions to center the object}

Encode state: $z_s \gets \text{concat}(E^{\text{img}}_{\phi}(x_{\text{img}}), E^{\text{inst}}_{\phi}(x_{\text{instr}}))$\;

\While{object not centered (within $\epsilon$ pixels)}{
    \ForEach{$a \in \mathcal{A}$}{    
        Encode action: $z_a \gets E^{\text{act}}_{\phi}(a)$\;
        Predict delta: $\Delta_\theta \gets f_\theta(z_s, z_a)$\;
        Predict next state: $\hat{z}_{s'} \gets z_s + \Delta_\theta$\;
        Compute distance to goal: $d_a \gets D(\hat{z}_{s'}, z^*)$\;
    }
    Select action: $a^* \gets \arg\min_a d_a$\;
    Execute $a^*$ on the drone\;
    Update state embedding: $z_s \gets \text{concat}(E^{\text{img}}_{\phi}(\text{new image}), E^{\text{inst}}_{\phi}(x_{\text{instr}}))$\;
}
\end{algorithm}

\subsection{Prompts for LLM}\label{app:prompt}

In this section, we provide the full prompts used for both multi-modal LLMs to generate the required actions to center the object in the image.

\begin{figure}[h]
\noindent
\begin{tcolorbox}[promptbox, title=Output: Prompt for raw images]
    Analyze this image taken by a drone for a gauge and predict the direction the drone should move so that the gauge is centered in the image. \\
    Assume that the gauge is within the field of view of the drone, and the image is taken from the front of the drone.\\
    The drone is able to move left, right, up, down, or not move at all.\\
    The response must include both horizontal and vertical movement directions.\\
    The response must be in the following format: 
    ["move x": "left", "move z": "up"], ["move x": "right", "move z": "down"], ["move x": "none", "move z": "none"]
\label{prompt}
\end{tcolorbox}
\end{figure}

\begin{figure}[h]
\noindent
\begin{tcolorbox}[promptbox, title=Output: Prompt for annotated images]
    Analyze this image taken by a drone for a gauge and predict the direction the drone should move so that the gauge is centered in the image. \\
    This image contains bounding box annotations of a gauge as well as a dot to indicate the center of the image.\\
    Assume that the gauge is within the field of view of the drone, and the image is taken from the front of the drone.\\
    The drone is able to move left, right, up, down, or not move at all.\\
    The response must include both horizontal and vertical movement directions.\\
    The response must be in the following format: 
    ["move x": "left", "move z": "up"], ["move x": "right", "move z": "down"], ["move x": "none", "move z": "none"]
\label{annotated_prompt}
\end{tcolorbox}
\end{figure}

\subsection{Model hyperparameters}\label{app:hyperparam}

This section tabulates the specific details on the model architecture and hyperparameters used to train latent dynamics model for this work. 

\begin{table}[h]
\small
\centering
\caption{Model Architecture Components and Details}
\begin{tabular}{|l|p{7cm}|p{3cm}|}
\hline
\textbf{Component} & \textbf{Architecture / Details} & \textbf{Dimensions / Config} \\
\hline
\multirow{5}{*}{Image Embedder} 
& ResNet-18 style with residual connections & Input: 3×224×224; Output: 128 \\
& Stem: Conv2d(7×7, stride=2) + MaxPool & - \\
& 4 residual layers: [64,128,256,512] & - \\
& Residual block: 3×3 Conv + BatchNorm + ReLU & - \\
& Global average pooling + linear projection & - \\
\hline
\multirow{4}{*}{Text Embedder} 
& Token embedding + positional encoding & Embedding: 128; Vocabulary: 5000 \\
& DistilBERT tokenizer & - \\
& 2 transformer encoder layers; Multi-head attention (4 heads) & Max length: 50 \\
& Feed-forward dim: 256; Dropout: 0.1 & - \\
\hline
\multirow{3}{*}{Action Embedder} 
& Separate embedding tables for X/Z axes & 16-dim per action → 32-dim total \\
& 3 actions per axis (left/right/none, up/down/none) & - \\
& Concatenated output & - \\
\hline
\multirow{3}{*}{Latent Dynamics Model} 
& MLP: 160 → 128 → 128 → 128 & Predicts state transformation in latent space \\
& Input: state embedding + action embedding & - \\
& ReLU activation; Dropout: 0.2 & - \\
\hline
\multirow{2}{*}{Global Action Embeddings} 
& Learnable 3×128 parameter matrices per axis & 3×128 per axis \\
& Provides consistent action semantics & - \\
\hline
\end{tabular}
\end{table}

\begin{table}[h]
\small
\centering
\caption{Training Hyperparameters and Loss Configuration}
\begin{tabular}{|l|l|l|}
\hline
\textbf{Category} & \textbf{Parameter} & \textbf{Value} \\
\hline
\multirow{4}{*}{Optimization} & Optimizer & Adam \\
& Learning Rate & 3e-4 \\
& Batch Size & 64 \\
& Epochs & 50 \\
\hline
\multirow{3}{*}{Data Processing} & Image Size & 224×224 \\
& Validation Split & 0.2 \\
& Sequence Length & 50 tokens \\
\hline
\multirow{2}{*}{Loss Hyperparameters} & Margin & 0.1 \\
& Temperature & 1.0 \\
\hline
\multirow{4}{*}{Augmentation} & Color Jitter & brightness, contrast, saturation, hue \\
& Random Affine & shear \\
& Gaussian Noise & N/A \\
& Instruction Variations & 30 paraphrased versions \\
\hline
\end{tabular}
\end{table}

\end{document}